# Deep Transform: Error Correction via Probabilistic Re-Synthesis

Andrew J.R. Simpson [#1]

[#] Centre for Vision, Speech and Signal Processing, Surrey University
Surrey, UK
[1] Andrew.Simpson@Surrey.ac.uk

*Abstract*—**Errors in data are usually unwelcome and so some means to correct them is useful. However, it is difficult to define, detect or correct errors in an unsupervised way. Here, we train a deep neural network to re-synthesize its inputs at its output layer for a given class of data. We then exploit the fact that this abstract transformation, which we call a *deep transform* (DT), inherently rejects information (errors) existing outside of the abstract feature space. Using the DT to perform probabilistic re-synthesis, we demonstrate the recovery of data that has been subject to extreme degradation.**

## I. Introduction

There are many situations in which errors are introduced into data. For example, errors may be introduced during transmission or storage of data. These errors are usually unwelcome because they make the data less useful by distorting or losing information. Therefore, some means to correct the errors is useful. Even more useful would be some unsupervised means to correct errors that would exploit prior knowledge about the scope of what is represented by the data.

Without prior knowledge, it is impossible to know what component of some given data might be error. For example, a list of binary digits might contain errors but since the errors are also binary digits it is impossible to know whether a given digit is an error or not. However, if there is a known degree of abstraction, the situation is different. For example, given a list of phone numbers comprising sequences of integers, it would be possible to detect those integers which are greater than 9 or less than 0. Further knowledge, e.g., of country codes, would allow more abstract errors to be detected. Thus, the greater the degree of known abstraction, the more powerful the detection of error will be. In this case, we may think of the abstractions as *features*.

Deep neural networks [1], [2], [3], [4], [5] (DNN) learn abstract feature representations from data [6]. A DNN that is trained to replicate its inputs at its output layer provides a one-to-one mapping and is known as an autoencoder [2]. Autoencoders implicitly learn both abstract encoding and decoding (i.e., synthesis). Once trained, we may think of the autoencoder as an abstract transformation device – a *deep transform* (DT) – that embodies abstract knowledge of the data. The DT is therefore capable of transforming new data through the learned abstract feature space, constituting an abstract filter. Critically, if the error is to pass through the DT (i.e., by being re-synthesized at the output) it must be encoded with the same abstract coding scheme. Thus, in principle, the DT may be employed as an abstract filter to reject error. Furthermore, the DT may therefore also be used to synthesize probabilistic corrections for errors in an unsupervised manner.

In this paper, we trained a DNN autoencoder on data that were free of errors and then used the resulting model as a transformation device (i.e., a DT). We then used this DT to probabilistically re-synthesize data that had been degraded by having a random proportion replaced with random data. We then trained a classifier DNN on the same un-degraded training data and demonstrate the recovery of separate test data from serious error using the DT to perform probabilistic re-synthesis.

## II. Method

By way of case study, we consider the problem of image classification for images that have been arbitrarily degraded by errors. We chose the well-known MNIST hand-written digit classification problem [3], [1], [4] and trained a typical feed-forward DNN to classify the digits. We then quantified the sensitivity of the trained classifier to error by testing the trained model on data that had been subjected to various degrees of nonlinear (non-additive) error. Next, to obtain a DT, we trained an autoencoder DNN on the same hand-written digit images. We then used the DT to re-synthesize each degraded image a number of times, each time with further random errors added, and averaged the result. Finally, to capture the degree of recovery from error, we tested the same classifier model on the re-synthesized (i.e., corrected) images.

For the input layer to our networks we unpacked the images of 28x28 pixels into vectors of length 784 (28x28 = 784). Pixel intensities were normalized to zero mean. We constructed two feed-forward DNNs; an autoencoder for use as DT and a classifier. Both networks used biased sigmoid activation functions [6]. The classifier network was of size 784x100x10 units, with a 10-unit softmax output layer

(corresponding to the 10-way classification problem). The autoencoder was of size 784x1500x784 units and featured a sigmoidal output layer. The autoencoder was trained to replicate its own inputs at its output layer.

Both models were independently trained on the 60,000 training examples from the MNIST dataset [4] using stochastic gradient descent (SGD). The models were trained for 10 iterations each. Each iteration of SGD consisted of a complete sweep of the entire training data set. Models were trained without dropout.

*Degradation.* In order to degrade the images for the testing process, a certain proportion of pixels from each of the 10,000 separate test images was randomly chosen and replaced with a random intensity selected from a distribution of equivalent mean and standard deviation to the images. Images were degraded at levels between 0 and 100% (of the pixels) at intervals of 10. Fig. 1a provides examples of digit images and images degraded to a level of 75% replacement. Note that this is not the same as additive noise.

*Probabilistic Re-Synthesis via DT.* Each degraded image was transformed using the autoencoder 100 times. Prior to each of the 100 separate transforms, 50% of the pixels of the degraded image (chosen at random) were replaced with random intensity values. I.e., each time a degraded image was further degraded, but these secondary degradations were different for each transform. The activation of the output layer of the autoencoder was taken as re-synthesized image. The resulting distribution, of 100 re-synthesized instances of each degraded image, was then averaged to provide an error-corrected image. In order to account for neurons in the output layer that were invariantly active, all error-corrected images (i.e., of the entire test set) were then averaged and the result subtracted from each image.

The same classifier was then separately tested on the both the degraded images and the error corrected images (without any additional training). Note that at 100% degradation the entire image is replaced with random noise. Hence, classifier accuracy is bounded by a ceiling of 10% chance level.

### III. RESULTS

Fig. 1a provides example images from the MNIST dataset in their original form (left), after 75% degradation (middle) and after recovery via DT probabilistic re-synthesis (right). Even with exact knowledge of what is represented in the degraded images, it is difficult or impossible to visually identify the digits. In contrast, the recovered images are clear and true to the original classification. However, note that the recovered images are not identical to the originals in shape and form. Indeed, it looks rather like the original images have been enhanced, suggesting the action of abstract re-synthesis.

This impressive recovery is reflected in the classification accuracy difference before and after DT probabilistic re-synthesis. Fig. 1b plots classification error as a function of degradation [%] for the classifier without DT probabilistic re-synthesis and for the classifier with DT probabilistic re-synthesis. In the degradation range between 20 and 90 % there is a large improvement for the classifier using DT probabilistic re-synthesis. This means that the classifier is far more robust to error when the degraded data are pre-processed with the DT probabilistic re-synthesis. Indeed, at a degradation rate of 70%, the classifier has an error rate of nearly 54% whereas with the error-corrected data the same classifier has an error rate of around 18%. In summary, we have demonstrated recovery from errors via DT probabilistic re-synthesis.

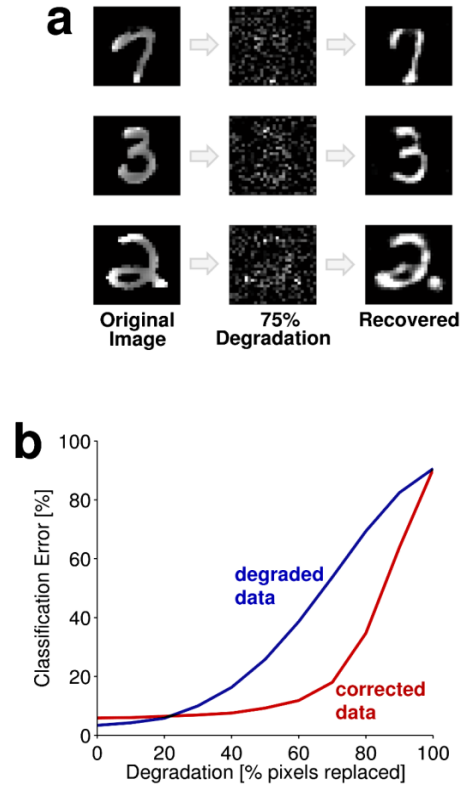

**Fig. 1. Probabilistic re-synthesis via deep transform. a** shows example images from the MNIST dataset (left) with highly degraded instances (middle) and images corrected by DT probabilistic re-synthesis (right). **b** Classification error in the test set as a function of degradation for degraded images (blue) and corrected images (red).

In order to provide a more practical demonstration of recovery from more typical errors (e.g., from network packet loss during broadcast or transmission), we used the same DT probabilistic re-synthesis to correct images that had been subject to large (~12% of the pixels) continuous regions being removed (zeroed) from a central (i.e., important) region in the image. Fig. 2a shows the original images, the images with missing regions and the images recovered using DT probabilistic re-synthesis. The resulting large gaps in the flowing strokes of each hand-written digit are incongruous. In the recovered images, these gaps have been replaced with smooth, flowing lines that are consistent with the hand-written qualities of the original. In the case of the '5', the upper vertical line has actually been adjusted towards a more archetypal '5'. Thus, our DT probabilistic re-synthesis not only corrects errors at the level of missing pixels but also appears to correct stylistic mistakes of the original writer. In other words, the DT probabilistic re-synthesis appears to be capable of correcting abstract errors introduced by the original writer.

To further illustrate this facility for abstract error correction, we took characters (not digits) from the Verdana Font family, rendered at the same 28x28 resolution, and performed the same DT probabilistic re-synthesis. The resulting images, and the original images, are shown in Fig. 2b. While there is no *a-priori* notion of error in the Verdana font, the abstract error (from the perspective of the model trained on hand-written digits) is a lack of hand-written qualities. The re-synthesized images have a hand-written quality that suggests 'correction' of the type-set style of the font.

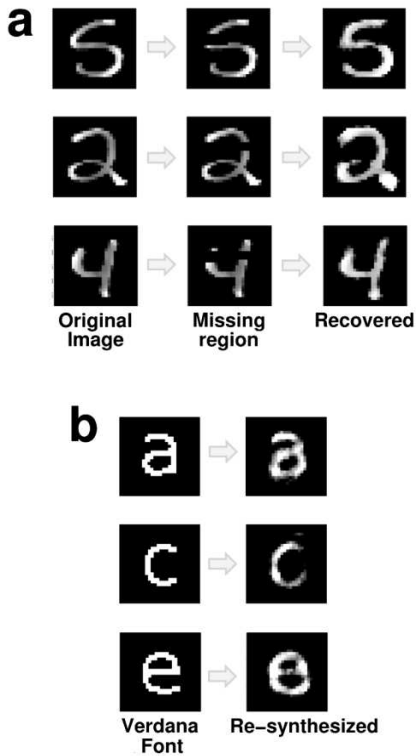

**Fig. 2. Practical illustration of probabilistic re-synthesis via deep transform. a** shows example images from the MNIST dataset (left) with regions missing (middle) and images corrected by DT probabilistic re-synthesis (right). **b** shows letters from the Verdana Font 'corrected' by DT probabilistic re-synthesis.

## IV. DISCUSSION AND CONCLUSION

In this paper, we have interpreted the DNN autoencoder as an abstract transformation device. We call this a deep transform because it is an abstract transformation learned by a deep neural network [6]. We have shown that this DT can be used for probabilistic re-synthesis and that this procedure allows data to be recovered from extreme error. This point was illustrated by contrasting the classification performance, for the same trained classifier, when making predictions about the degraded and corrected data. Indeed, this seems more remarkable given that the classifier was not re-trained to either account for, or to exploit, any training data that had been re-synthesized.

We have also demonstrated, at a more intuitive level, that DT probabilistic re-synthesis can correct abstract errors and even affect stylistic changes. Thus, it would appear that there is great potential for DT probabilistic re-synthesis to perform similarly in various related tasks. For example, we have shown that typeset font characters can be transformed into characters that have hand-written qualities. This demonstrates that abstract qualities of the hand-written digits learned (by the autoencoder) may be applied to re-synthesize letters of the alphabet that were never encountered in the training stage.

More generally, the outcome of the DT probabilistic re-synthesis demonstrated here resembles the perceptual error correction facilities of the brain. Indeed, our demonstration of replacement of large regions of missing digit appears consistent with the broadly equivalent auditory perceptual phenomenon of 'missing phoneme replacement'. Hence, our findings may provide insight into the exceptionally robust perceptual system of the brain and its illusions.

ACKNOWLEDGMENT

AJRS did this work on the weekends and was supported by his wife and children.